\documentclass[letterpaper, 10 pt, conference]{ieeeconf} 
\IEEEoverridecommandlockouts

\usepackage{amsmath}
\usepackage{amssymb}
\usepackage{amsthm}
\usepackage{mathtools}
\usepackage{amsfonts}

\PassOptionsToPackage{table}{xcolor}
\usepackage{xcolor}
\usepackage{color}
\usepackage{colortbl}

\usepackage{graphicx}
\usepackage{wrapfig}
\usepackage[export]{adjustbox}
\usepackage{subcaption}
\usepackage[format=plain,font=small,labelfont=bf,labelsep=period]{caption}
\usepackage{sidecap} 
\usepackage{float}
\usepackage[format=plain,font=footnotesize,labelfont=bf,labelsep=period]{caption}

\usepackage{booktabs}       
\usepackage{multirow}
\usepackage{makecell}
\usepackage{hhline}
\usepackage{array}

\let\labelindent\relax
\usepackage{enumitem}

\usepackage{algorithm}
\usepackage{algpseudocode}
\usepackage{circledsteps}

\usepackage{nicefrac}       
\usepackage{bm}
\usepackage{cancel}
\usepackage{oubraces}
\usepackage{xfrac}
\usepackage{upgreek}

\usepackage[c2 , nocomma]{optidef}

\usepackage{microtype}      
\usepackage{xspace}
\usepackage{textcomp}
\usepackage{dirtytalk}
\usepackage{stackengine}

\usepackage[space]{grffile} 
\usepackage{moreverb, url}
\usepackage{verbatim}
\usepackage{cite}
\usepackage{hyperref}

\usepackage{comment}
\usepackage{pifont}
\usepackage{siunitx}
\usepackage{lipsum}
\usepackage{dblfloatfix}

\usepackage{times}
\usepackage{setspace}
\usepackage[utf8]{inputenc}
\usepackage[T1]{fontenc}
\newcolumntype{U}{S} 

\definecolor{blond}{rgb}{0.98, 0.98, 0.82}
\definecolor{blue}{RGB}{38,38,134}
\definecolor{darkblue}{RGB}{0,0,102}
\definecolor{lightblue}{RGB}{77,77,148}
\definecolor{gold}{RGB}{234, 170, 0}
\definecolor{metallic_gold}{RGB}{139, 111, 78}

\newtheorem{theorem}{Theorem}

\theoremstyle{definition}
\newtheorem{definition}{Definition}
\theoremstyle{remark}

\theoremstyle{definition}

\newcommand{\R}{\mathbb{R}}
\newcommand{\C}{\mathcal{C}}

\newcommand{\mb}[1]{\mathbf{ #1 }}

\newcommand{\algname}{D-SafeMPC\xspace}

\def\BibTeX{{\rm B\kern-.05em{\sc i\kern-.025em b}\kern-.08em
    T\kern-.1667em\lower.7ex\hbox{E}\kern-.125emX}}

\IEEEoverridecommandlockouts  
\begin{document}
\begin{spacing}{0.96}

\title{ 
\textbf{D-SafeMPC: Diffusion-Driven Safe Model Predictive Control \\ with Discrete-Time Control Barrier Functions}
}

\author{
Erdi Sayar$^{1}$, Ersin Daş$^{2}$, Joel W. Burdick$^{3}$, Alois Knoll$^{4}$, and Erdal Kayacan$^{1}$
\thanks{*This work was partially supported by the Horizon Europe Grant Agreement No.~101136056.}
\thanks{
$^{1}$ E. Sayar and E. Kayacan are with the Department of Electrical Engineering
and Information Technology, Paderborn University, 33098 Paderborn, Germany, \texttt{\{erdi.sayar, erdal.kayacan\}@uni-paderborn.de}.
}
\thanks{$^{2}$ E. Da\c{s} is with the Department of Mechanical, Materials, and Aerospace Engineering, Illinois Institute of Technology, Chicago, IL 60616, USA, {\texttt{edas2@illinoistech.edu}}.
}
\thanks{$^{3}$ J. W. Burdick is with the Mechanical and Civil Engineering, California Institute of Technology, Pasadena, CA 91125, USA, \texttt{jburdick@caltech.edu}.}%
\thanks{$^{4}$ A. Knoll is with the School of Computation, Information, and Technology, Technical University of Munich, 80333 Munich, Germany, \texttt{k@tum.de}.}%
}

\maketitle
\thispagestyle{empty}
\pagestyle{plain}

\begin{abstract}
A key limitation on the use of diffusion models in robotic planning is their inability to inherently enforce safety or dynamical constraints, which often results in physically infeasible or unsafe outputs. Hybrid approaches that employ model predictive control (MPC) to address this problem can be unstable, as poor trajectory initializations from the diffusion model prevent the MPC from converging to a safe and feasible solution. To overcome these challenges, we propose \algname, which enhances the interaction between diffusion and control. Our method guides the reverse diffusion process with control barrier functions (CBFs) and control Lyapunov functions (CLFs) and employs an iterative-projection scheme where an MPC refines the trajectory at each denoising step. This steers sampling toward safe, goal-directed regions and provides reliable MPC warm starts. In simulations on a Franka manipulator across four scenarios (one static-obstacle and three dynamic-obstacle settings) and in a sim-to-real experiment on a physical Franka robot, \algname improves safety, task success rates, and planning efficiency over state-of-the-art baselines.
To facilitate reproducibility, our source code and experimental configurations are available in a repository at ~\url{https://github.com/erdiphd/D-SafeMPC}

\end{abstract}


\begin{figure*}[t]
  \centering
  \includegraphics[width=1\linewidth, , trim={0 9.0cm 5cm 0}, clip]{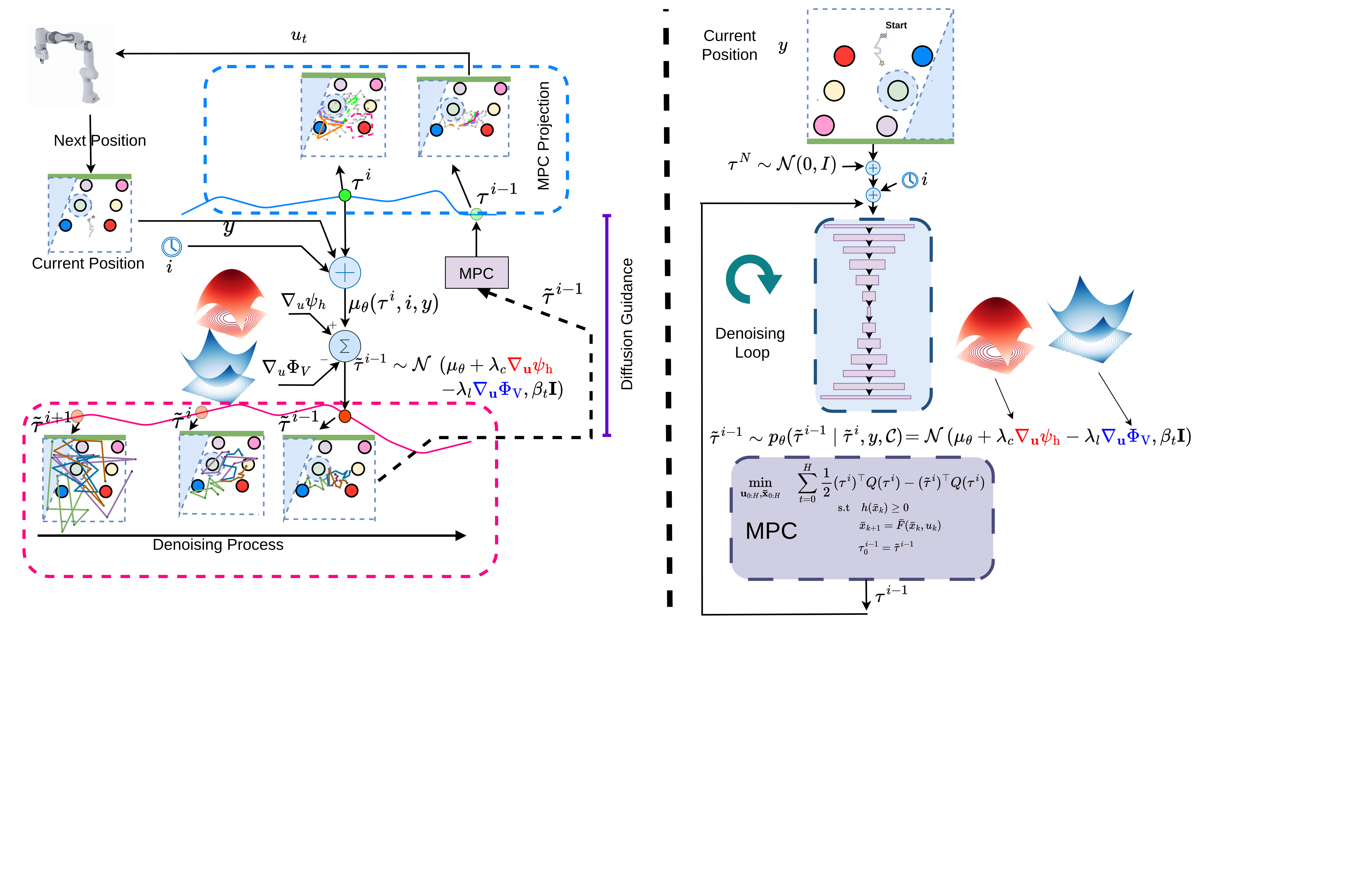}
  \caption{Overview of the proposed diffusion-based trajectory planning framework, \algname. The left side illustrates the evolution of trajectories during the denoising and projection process. At each denoising step, the diffusion model computes the mean $\mu_{\theta}(\tau^i, i, y)$ based on the projected trajectory from MPC $(\tau^i)$, the current denoising step $(i)$ and condition information $(y)$. This mean is then adjusted using gradients from the discrete-time CBF and CLF, denoted as $\nabla_{\mathbf{u}}\psi_{h}$ and $\nabla_{\mathbf{u}}\Phi_{V}$, respectively. The adjusted mean is then sampled to produce a trajectory $\tilde{\tau}^{i-1}$, which serves as initialization for the MPC. The MPC projects this trajectory onto a safe and dynamically feasible set $\tau^{i-1}$, and the projected trajectory initializes the next denoising iteration. This cycle continues until the final timestep, where the control action $u_t$ from the projected trajectory is applied to advance the robot to the next state $x_t$. This updated state then provides conditional input for generating the subsequent trajectory. The right side details the denoising process: the reverse diffusion loop begins with pure Gaussian noise conditioned on the current state $x_t$. Throughout denoising, the discrete-time CBF and CLF gradients guide the trajectory generation, steering it away from obstacles and toward the goal.}
  \label{fig:model_architecture}
    \vspace{-2mm}
\end{figure*}

\section{INTRODUCTION}
Diffusion models \cite{NEURIPS2020_4c5bcfec} have established state-of-the-art performance in generative tasks such as image and video synthesis. In robotics, they demonstrate impressive capabilities in trajectory planning, where they learn complex policies from demonstrations \cite{janner2022planning, chi2025diffusionpolicy, carvalho2023motion, carvalho2025motion}. These models excel at capturing multimodal behaviors and generating high-dimensional trajectories holistically, which mitigates the compounding errors common in sequential methods and improves long-term performance. However, despite their success with large datasets, diffusion models lack an intrinsic mechanism to enforce explicit constraints, such as system dynamics, obstacle avoidance, and safety margins \cite{xiao2023safediffuser}. As a result, the trajectories produced via their iterative stochastic denoising process frequently violate the physical constraints of the system \cite{kurtz2025equality}.
To address this limitation, \cite{mizuta2024cobl} combines a diffusion model with control barrier functions (CBFs) \cite{ames2017control} and control Lyapunov functions (CLFs) \cite{AmesCooganEtAl2019} to avoid obstacles while reaching a target set during the reverse diffusion process. Although this approach guides the planner, it only acts as a form of guidance, pushing generated trajectories toward safe and goal-directed regions. 
In other words, CBFs and CLFs improve the probabilistic safety of the planned trajectories, but no hard safety constraints are enforced on the final trajectory, which may still violate safety or dynamic constraints. 

To enforce such hard constraints, an alternative strategy is to leverage model predictive control (MPC) \cite{eren2017mpc}, an optimization-based control framework that computes control actions by repeatedly solving a finite-horizon optimal control problem subject to explicit system constraints. Owing to its ability to systematically handle complex dynamics and safety requirements, MPC has been widely adopted across robotic domains \cite{10406329,10916556}.
Recent approaches \cite{romer2025dpcc, zhou2024diffusion} address this limitation by employing a hybrid framework that couples a diffusion model with MPC. This process involves generating an initial trajectory with the diffusion model, using MPC to project it onto a safe and dynamically feasible set, and then feeding the corrected trajectory back to guide the next denoising step \cite{romer2025dpcc}. This iterative cycle enables MPC to continually enforce safe operational boundaries.
However, this hybrid strategy has a critical limitation: a poor initial trajectory from the diffusion model can prevent the MPC from converging to a feasible solution, leading to instability or failure in complex environments.

To address these challenges, we introduce \algname, a novel framework that seamlessly integrates diffusion-based trajectory generation with MPC leveraging CBFs and CLFs.
In each denoising iteration, we guide the reverse diffusion process by steering the generated trajectory samples toward safe, goal-directed regions. As aforementioned, this guidance steers the denoising process dynamically toward safe and goal-reaching trajectories without requiring retraining of the diffusion model, thereby rendering them probabilistically safer. However, this guidance alone does not enforce hard constraints such as safety or dynamic feasibility. Therefore, we refine these guided trajectories using MPC with explicit safety and dynamic constraints at each denoising step.
The integration of CBFs and CLFs provides MPC with well-initialized trajectories, enabling more stable and efficient convergence to safe and feasible solutions within the allocated computational budget. The refined trajectories from MPC are then used to initialize the subsequent denoising iteration. This iterative cycle continues until the final timestep, at which point the control action from the converged trajectory is applied to the robot to advance to the next state.

The main contributions of this work are:
\begin{itemize}
    \item We introduce a novel framework that leverages discrete-time CBFs and CLFs to guide the reverse diffusion process, steering trajectory generation toward safer, goal-oriented regions and improving MPC initialization.
    \item The MPC enforces strict safety and dynamic constraints on the probabilistically safe trajectories generated via discrete-time CBF and CLF guidance, ensuring that the resulting trajectory remains within a defined safer set and adheres to the dynamic constraints.
    \item Empirical validation on a Franka robot manipulator in static and dynamic obstacle scenarios, demonstrating superior safety, success rates, and computational efficiency compared to state-of-the-art baselines.
\end{itemize}

\section{Related work} \label{sec:related_work}
Diffuser \cite{janner2022planning} is an offline-trained diffusion model that learns to denoise entire state–action trajectories by treating them as single-channel images. Using pre-collected trajectory datasets, the model is trained to reconstruct clean trajectories from noise through iterative temporal denoising. During inference, Diffuser produces full-length trajectories by starting from Gaussian noise and applying two mechanisms: reward-guided sampling, where a learned critic steers the denoising process toward trajectories with higher expected return, and goal-conditioned inpainting, where specified start and end states are enforced. Despite its flexibility in generating goal-directed motion, this planning approach does not guarantee safety—there is no formal mechanism to prevent unsafe states during generation. To address this safety issue, CoBL-Diffusion \cite{mizuta2024cobl} combines a diffusion-based planner with CBFs and CLFs to enable safe navigation in dynamic environments. In this approach, CBFs and CLFs guide the model away from obstacles and toward goals. However, this guidance acts as a \textit{soft} preference rather than a \textit{hard} constraint, meaning the generated trajectories might still violate safety or dynamic constraints. Similarly, other work has focused on improving physical realism by aligning the diffusion model's sampling path with a traditional optimization trajectory, guiding the generation toward more valid solutions \cite{giannone2023aligning}. 

In contrast to these soft guidance approaches, an alternative technique is to enforce safety and goal-reaching requirements by projecting diffusion-generated trajectories onto the safe sets. The method proposed in \cite{romer2024safe} projects a noisy trajectory onto a constraint set while considering the system dynamics.
Diffusion predictive control with constraints (DPCC) \cite{romer2025dpcc} employs a diffusion model to generate horizon-based trajectories. These trajectories serve as an initial guess for an MPC, which then projects them onto a safe and dynamically feasible set by enforcing constraints on obstacle avoidance, safe operational regions, and system dynamics. The resulting projected trajectory initializes the next denoising step in the diffusion model, creating an iterative cycle in which the MPC continually refines the trajectory to ensure safety.
However, if the initial trajectory generated by the diffusion model is poor, the MPC may fail to find a feasible solution within its allocated iterations, which can pose a significant safety risk. This limitation becomes more pronounced in challenging scenarios, such as when dynamic obstacles are present or when obstacles intersect the paths of the expert trajectories used for training. Our method addresses this by guiding the diffusion output with CBFs and CLFs, which provides a high-quality initial guess for the MPC and enables convergence.

\section{Background} \label{sec:background}

Since the robot dynamics are typically continuous-time and nonlinear, we start by considering continuous-time control-affine nonlinear systems of the form:
\begin{align} 
\label{eq:affine-dynamics_c}
    \dot{\mb x } = \mb{f}(\mb{x}) + \mb{g}(\mb{x}) \mb{u}, \
    \mb{x} \in \mathcal{X} \subseteq \R^n, \
    \mb{u} \in \mathcal{U} \subseteq \R^m,
\end{align}
where the drift dynamics ${\mb f \!:\! \mathcal{X} \!\to\! \R^n}$ and the actuation matrix ${ \mb g \!:\! \mathcal{X} \!\to\! \R^{n \times m}}$ are locally Lipschitz continuous functions. We assume the set of admissible control inputs $\mathcal{U}$ is an $m$-dimensional compact convex polytope. Digital implementation of any controller introduces a sample-and-hold effect on the control:
\begin{equation}
\label{eq:ZOH}
\mb{u}(t) = \mb{u}_k, \quad \forall t \in [t_k, t_{k+1}),
\end{equation}
where $\mb{u}_k$ is constant between consecutive sampling instants $t_k$ and $ t_{k+1}$ with the time index $k \in \mathbb{N}_0$. We assume a uniform sampling strategy with a sampling interval $T$. As the system is piecewise locally Lipschitz continuous, it admits a unique trajectory. We can obtain a discrete-time nonlinear model of the continuous-time system \eqref{eq:affine-dynamics_c} under the sample-and-hold strategy \eqref{eq:ZOH} as
\begin{equation} 
\label{eq:affine-dynamics}
\!\!{\mb x }_{k+1} \! \!=\! \mb x_k  + \!\! \int^{t_{k+1}}_{t_k} \!\! \left ( \mb{f}(\mb x(\tau) \!) \!+\! \mb{g}(\mb x(\tau)) \mb{u}_k \!\right )  d\tau \!\triangleq\! \mb{F}(\mb{x}_k, \!\mb{u}_k ),
\end{equation}
where $\mb{F} : \mathcal{X} \times \mathcal{U} \!\to\! \mathcal{X}$ is the \textit{exact state discrete map}. If we use a one-step Euler discretization to approximate this map, we have ${{\mb x }_{k+1} \!=\! \mb{F}(\mb{x}_k, \mb{u}_k) \!=\! \mb x_k \!+\! \left ( \mb{f}(\mb{x}_k) \!+\! \mb{g}(\mb{x}_k) \mb{u}_k \right ) T} \!+\! \mathcal{O}(T^2) $. Throughout the paper, we denote $\mb{F}$ as the discrete-time model used for control and analysis. When necessary, $\mb{F}$ refers to the Euler approximation of the exact flow map.

\subsection{Discrete-Time CBFs and CLFs}
\label{subsec:CBF}
We consider dynamic obstacles in our safe control scenarios, and our control framework is discrete-time. We first define discrete-time control barrier functions (CBFs) \cite{agrawal2017, ahmadi2019safe}, which extend continuous-time CBFs \cite{ames2017control} to certify safety at sampling instants. 

We characterize safety through the notion of forward invariance. Let the time-invariant set ${\C }$ be the 0-superlevel set of a continuously differentiable function ${h\!:\! \mathcal{X} \!\to\! \mathbb{R}}$:
\begin{equation}
\label{eq:set}
    \C\!\triangleq\! \left\{ \mb x \!\in\! \mathcal{X} \!\subseteq\! \mathbb{R}^n \!:\! h( \mb x) \!\geq\! 0 \right\}.
\end{equation}
This set is said to be \textit{forward invariant} if for the discrete-time system \eqref{eq:affine-dynamics}, for any initial condition ${\mb x_0 \!\in\! \C}$, implies that ${\mb x_k \in \C}$, for all ${k \in \mathbb{N}_{0}}$. The closed-loop system ${\mb x_{k+1} = \mb{F}(\mb{x}_k, \!\mb k( \mb x_k) )}$, with a feedback controller ${\mb k \!:\! \mathcal{X} \!\to\! \mathcal{U}}$, ${\mb u \!=\! \mb k( \mb x)}$, is {\em safe} with respect to the \textit{safe set} $\C$, if $\C$ is forward invariant. To design controllers that guarantee such forward invariance properties, discrete-time control barrier functions (CBFs) provide a systematic framework.
\begin{definition}[Discrete-Time Control Barrier Function]
\label{def:cbf}
Let ${\C \!\subseteq\! \mathcal{X} }$ be the 0-superlevel set of a continuously differentiable function ${h\!:\! \mathcal{X} \!\to\! \mathbb{R}}$. The function $h$ is a \textit{discrete-time control barrier function} for system \eqref{eq:affine-dynamics} with respect to $\C$ if there exists a class-$\mathcal{K}$ function\footnote{A continuous function ${\alpha \!:\! [0, a ) \!\to\! \mathbb{R}_{\geq 0}}$, where ${a \!>\! 0}$, belongs to class-${\mathcal{K}}$ (${\alpha \!\in\! \mathcal{K}}$) if it is strictly monotonically increasing and ${\alpha(0) \!=\! 0}$.} ${\alpha \!\in\! \mathcal{K}}$ satisfying ${\alpha(r) \!<\! r, \ \forall r \!>\! 0}$ such that ${\forall \mb x_k \!\in\! \C}$:
\begin{equation*}
   \sup_{\mb u_k \in \mathcal{U}} \big [ h(\mb{F}(\mb{x}_k, \!\mb{u}_k )) - h(\mb x_k) \big ] 
   \!>\! -\alpha (h(\mb x_k)).
\end{equation*}
\end{definition}

We can derive formal safety guarantees using Definition~\ref{def:cbf}, with the help of the following theorem:
\begin{theorem} \label{thm: cbf}
If $h$ is a discrete-time CBF for \eqref{eq:affine-dynamics} on $\C$, then any feedback controller ${\mb k: \mathcal{X} \!\to\! \mathcal{U}}$, ${\mb u_k = \mb k( \mb x_k)}$ satisfying 
\begin{align} \label{eq: cbf_condition}
    h(\mb{F}(\mb{x}_k, \!\mb k(\mb{x}_k ))) - h(\mb x_k) 
   \!\geq\! -\alpha (h(\mb x_k))
\end{align}
for all ${\mb x_k \!\in\! \C}$ renders the set $\C$ forward invariant.
\end{theorem}

Similar to CBFs, CLFs encode stability objectives by requiring a Lyapunov candidate to decrease along closed-loop trajectories. Therefore, CLFs are useful for representing closed-loop objectives, such as stabilizing to a desired configuration or, through an appropriate choice of CLFs, converging to a target set. We specifically consider discrete-time exponential control Lyapunov functions (CLFs): 
\begin{definition}[Discrete-Time Control Lyapunov Function] \label{def:clf}
A continuously differentiable function $V(x)$, ${V \!:\! \mathcal{G} \!\subseteq\! \mathcal{X} \!\to\! \mathbb{R}_{\geq0}}$, is a \textit{discrete-time control Lyapunov function} for system \eqref{eq:affine-dynamics} if there exists a constant ${\sigma \in (0,1)}$ such that for all ${\mb x_k \!\in\! \mathcal{G}}$:
\begin{align*}
 \inf_{\mb u_k \in \mathcal{U}} \Big[ V(\mb{F}(\mb{x}_k, \mb{u}_k )) - V( \mb x_k) \Big]
   \leq -\sigma  V( \mb x_k).
\end{align*}
\end{definition}

\subsection{Diffusion Models}
\label{subsec:diffusion_models}
Diffusion models \cite{NEURIPS2020_4c5bcfec} express a probability distribution $p(\tau_0)$ through latent variables in the form  $p_{\theta}\left(\boldsymbol{\tau}_{0}\right)\triangleq\int p_{\theta}\left(\boldsymbol{\tau}_{0: N}\right) d \boldsymbol{\tau}_{1: N}$, where $\boldsymbol{\tau}_{1}, \ldots, \boldsymbol{\tau}_{N}$ are latent variables of the same dimensionality as the data $\boldsymbol{\tau}_{0} \sim p\left(\boldsymbol{\tau}_{0}\right)$.

These models are characterized by a forward and a reverse diffusion process. The forward diffusion process is the fixed approximate posterior $q\left(\boldsymbol{\tau}_{1: N} \mid \boldsymbol{\tau}_{0}\right)$, given by a Markov chain that, starting from  $\boldsymbol{\tau}_{0} \sim q\left(\boldsymbol{\tau}_{0}\right)$, perturbs the input data by gradually adding Gaussian noise in $N$ steps with  a predefined variance schedule $\beta_{1},\dots,\beta_{N}$ following the cosine noise schedule \cite{nichol2021improved}. This process is defined as
\begin{alignat}{2}
q\left(\boldsymbol{\tau}_{1: N} \mid \boldsymbol{\tau}_{0}\right) &\triangleq\prod_{i=1}^{N} q\left(\boldsymbol{\tau}_{i} \mid \boldsymbol{\tau}_{i-1}\right),\\
q\left(\boldsymbol{\tau}_{i} \mid \boldsymbol{\tau}_{i-1}\right)&\triangleq\mathcal{N}\left(\boldsymbol{\tau}_{i} ; \sqrt{1-\beta_{i}} \boldsymbol{\tau}_{i-1}, \beta_{i} \boldsymbol{I}\right). \nonumber
\end{alignat}

The reverse diffusion process aims to recover the original input data from the noisy (diffused) data.
A deep neural network with parameters $\theta$ is trained to progressively reverse the diffusion process step by step and approximates the joint distribution $p_{\theta}\left(\boldsymbol{\tau}_{0: N}\right)$. 
This process is modelled as \cite{NEURIPS2020_4c5bcfec}
\begin{alignat}{2}
p_{\theta}\left(\boldsymbol{\tau}_{0: N}\right) &\triangleq  p\left(\boldsymbol{\tau}_{N}\right) \prod_{i=1}^{N} p_{\theta}\left(\boldsymbol{\tau}_{i-1} \mid \boldsymbol{\tau}_{i}\right), \\
p_{\theta}\left(\boldsymbol{\tau}_{i-1} \mid \boldsymbol{\tau}_{i}\right)&\triangleq \mathcal{N}\left(\boldsymbol{\tau}_{i-1} ; \boldsymbol{\mu}_\theta\left(\boldsymbol{\tau}_i, i\right), \boldsymbol{\Sigma}_\theta\left(\boldsymbol{\tau}_i, i\right)\right), \nonumber
\end{alignat}
where $p\left(\boldsymbol{\tau}_{N}\right) = \mathcal{N}\left(\boldsymbol{\tau}_{N} ; \boldsymbol{0}, \boldsymbol{I}\right)$, $\boldsymbol{\mu}_\theta$ and $\boldsymbol{\Sigma}_\theta$ are the outputs of the neural network.
Following \cite{NEURIPS2020_4c5bcfec}, we set $\boldsymbol{\Sigma}_\theta = \sigma_i^{2} \boldsymbol{I} = \beta_{i} \boldsymbol{I}$, which is not learned but follows the cosine schedule, and the mean is defined as
\begin{equation}
    \boldsymbol{\mu}_\theta\left(\boldsymbol{\tau}_i, i\right) = \frac{1}{\sqrt{\alpha_i}} \left( \tau_{i} - \frac{\beta_i}{\sqrt{1 - \bar{\alpha}_i}} \epsilon_{\theta}(\tau_i, i) \right),
 \end{equation}
where ${\alpha}_i = 1 - \beta_i$ and $\bar{\alpha}_i=\prod_{s=1}^N \alpha_s$ is the cumulative product of the $\alpha$ terms. The model is optimized by maximizing the evidence lower bound (ELBO) \cite{Blei_2017}, $\mathbb{E}_{q}\left[\ln \frac{p_{\theta}\left(\boldsymbol{\tau}_{0: N}\right)}{q\left(\boldsymbol{\tau}_{1: N} \mid \boldsymbol{\tau}_{0}\right)}\right]$. The training objective can be simplified from the ELBO to a loss function \cite{NEURIPS2020_4c5bcfec}:
\begin{equation}
\mathcal{L}_d(\theta) = \mathbb{E}_{i, \epsilon ,\tau_0}
\bigg [ \Big\| \epsilon - \epsilon_{\theta} \big(\underbrace{\sqrt{\bar{\alpha}_i} \tau_0 + \sqrt{1 - \bar{\alpha}_i} \, \epsilon}_{\tau_i}, i \Big) \Big\|^2 \bigg],
\end{equation}
here, $\epsilon_{\theta}$ is a neural network that learns to predict the noise. Once trained, sampling data from Gaussian noise ${\boldsymbol{\tau}_{N} \!\sim\! p\left(\boldsymbol{\tau}_{N}\right)}$ and running it through the reverse diffusion process from ${i\!=\!N}$ to ${i\!=\!0}$ yields an approximation of the original data distribution.

\section{Trajectory Planning with Conditional Diffusion models}
\label{sec:proposed_model}

As planning and diffusion processes operate on distinct temporal values, we denote the diffusion iteration index with a superscript $ i \in \{0, \ldots, N \}$ and the planning trajectory index with a subscript  $t \in \{0, \ldots, T \}$. Following this convention, $\tau^i = (\mathbf{x}_0^i, \mathbf{u}_0^i, \dots, \mathbf{x}_H^i)$ represents the entire trajectory at diffusion step $i$ with a horizon $H$. However, an important distinction is that we denote the output of the guided diffusion model with $\tilde{\tau}^i$ and the output of the MPC projection with $\tau^i$.

We consider a robotic system operating in an environment with $N_J$ static or dynamic obstacles. The system is governed by the discrete-time, nonlinear dynamics given in \eqref{eq:affine-dynamics}. 
The state and control input of the $j$-th obstacle at timestep $t$ is denoted by $\mb{x}_{j,t}$ and $\mb{u}_{j,t}$, for ${j = 1, \dots, N_J}$.

Our objective is to control the robot from an initial state $\mb{x}_0$ to a goal ${g \!\in\! \mathcal{G}}$ within a maximum of $T$ time steps. Goal achievement is indicated by a binary function ${\phi \!:\! \mathcal{X} \!\times\! \mathcal{G} \!\to\! \{0, 1\}}$, where ${\phi(\mb{x}_T, g) \!=\! 1}$ signals success. In our examples, we use the D3IL dataset \cite{jia2024towards}, which contains teleoperated demonstrations with a 7-DoF Franka Emika Panda robot in MuJoCo \cite{todorov2012mujoco}. This dataset, ${\mathcal{D} \!=\! \{\prescript{(m)}{}{\tau_e}\}_{m=1}^M}$, consists of $M$ successful expert trajectories of the form ${\prescript{(m)}{}{\tau_e} \!=\! (\mb{\prescript{(m)}{}x}_0, \mb{\prescript{(m)}{}u}_0, \dots, \mb{\prescript{(m)}{}x}_T)}$, each associated with a goal $\prescript{(m)}{}g$ such that ${\phi(\mb{\prescript{(m)}{}x}_T, \prescript{(m)}{}g) \!=\! 1}$. The expert is modeled as an unknown stochastic policy $\pi_e$, where actions are sampled as ${\mb{u}_t \!\sim\! \pi_e(\cdot \mid \mb{x}_t, g)}$.
We aim to learn a goal-conditioned stochastic control policy $\pi$ from the expert dataset $\mathcal{D}$ using a conditional diffusion model. 
As recent work \cite{janner2022planning} has demonstrated that conditional diffusion models can effectively learn complex robotic control policies from offline data, we adopt this method. Our diffusion model is trained to approximate the conditional distribution of trajectories $p_{\theta}(\tau \mid y)$ where $y$ is the conditioning information, such as the current state $\mb{x}_t$ and a desired goal $g$.

During the training process, the diffusion model learns from offline trajectory data collected in obstacle-free environments, implicitly encoding dynamic constraints from the demonstration data. However, during inference, the environment may contain obstacles and unsafe regions not present in the training data. While the diffusion model can generate trajectories with horizon $H$ conditioned on $y$ that respect the learned dynamics, the stochastic nature of the reverse diffusion process may produce trajectories that violate safety constraints or dynamic feasibility.
To address these safety concerns, we define a safe set, denoted by $\mathcal{C}$, using a CBF, $h(\mb x)$. This set consists of all the states $\mb x$ for which the CBF is non-negative, as formally defined by the set in \eqref{eq:set}.

We can incorporate these constraints into the diffusion architecture and train the model in environments with dynamic obstacles. However, learning the distribution ${p_{\theta}(\tilde{\tau}^{i - 1} \!\mid\! \tilde{\tau}^{i}, y, \mathcal{C})}$ requires extensive offline demonstrations that satisfy the safety constraints. Furthermore, since $\mathcal{C}$ is task-specific, the trained model might fail to generalize during inference for unseen scenarios.

To overcome these challenges, we propose a more flexible approach: incorporating the safety constraints as soft constraints during the denoising process. We achieve this using classifier guidance \cite{NEURIPS2021_49ad23d1}, which dynamically steers trajectory generation toward safe regions without requiring model retraining. Specifically, we model the likelihood of a trajectory $\tilde{\tau}^{i-1}$ belonging to the safe set $\mathcal{C}$ using a Gibbs-Boltzmann distribution:
\begin{equation}
p(\tilde{\tau}^{i-1} \in \mathcal{C}) \propto \exp\left( \frac{h(\tilde{\tau}^{i-1})}{\mathcal{T}} \right),
\end{equation}
where $\mathcal{T}$ is a temperature parameter, and trajectories with higher (safer) $h$ values are assigned a higher probability, effectively directing the reverse diffusion process toward desired outcomes.

Our objective is thus to sample from the posterior $p(\tau^0 \mid y,\mathcal{C})$, expressed as
\begin{equation}
p(\tau^{0}\mid y,\mathcal{C})=\int p(\tau^{N}\mid y,\mathcal{C})
\prod_{i=1}^{N} p(\tau^{i-1}\mid \tau^{i},\, i,\,y, \mathcal{C}) \,\mathrm{d}\tau_{1:N}
\end{equation}
with $p(\tau^{N}\mid y, \mathcal{C}) = \mathcal{N}( \mathbf{0}, \mathbf{I})$.
Sampling from this posterior involves iteratively drawing from the safety-conditioned distribution via Bayes' rule:
\begin{equation}
  p(\tilde{\tau}^{i - 1} \mid \tilde{\tau}^{i}, i, y, \mathcal{C})  = Z \cdot p(\tilde{\tau}^{i-1} \mid \tilde{\tau}^i, i, y) \cdot p(\mathcal{C} \mid \tilde{\tau}^{i-1}),
\end{equation}
where $Z$ is a normalizing constant. Because $Z$ is typically intractable to compute, it is not possible to sample from this distribution directly \cite{NEURIPS2021_49ad23d1}. 
Therefore, the safety-conditioned distribution $p(\tilde{\tau}^{i - 1} \mid \tilde{\tau}^{i}, i, y, \mathcal{C})$ has no closed form for sampling.
Taking the logarithm of this safety-conditioned posterior yields
\begin{equation}
  \log p(\tilde{\tau}^{i - 1} \!\mid\! \tilde{\tau}^{i}, i, y, \mathcal{C}) \!\approx\! \log p(\tilde{\tau}^{i-1} \!\mid\! \tilde{\tau}^i, i, y) \!+\! \log p(\mathcal{C} \mid \tilde{\tau}^{i-1}),
\end{equation}
where $p(\tilde{\tau}^{i-1} \!\mid\! \tilde{\tau}^i, i, y) \approx p_{\theta}(\tilde{\tau}^{i-1} \!\mid\! \tilde{\tau}^i, i, y)$ denotes the learned denoising model. This decomposition separates the learned diffusion denoising from the safety guidance, and the learned denoising model is given by
\begin{align}
  &\log p_{\theta}(\tilde{\tau}^{i - 1} \mid \tilde{\tau}^{i}, i, y)  = \log \mathcal{N}(\mu_{\theta}, \beta_i \mathbf{I}) \\ \nonumber
  &~~~~~~~~~~~~~~~~= -\frac{1}{2}  \! \left(\tilde{\tau}^{i-1} \!-\! \mu_{\theta} \right)^\top \!\!\left(\!\frac{\mathbf{I}}{\beta_i} \! \right) \left(\tilde{\tau}^{i-1} \!-\! \mu_{\theta}\right) \!+\! c_{1},
  \end{align}
where $c_1$ is a constant and \(\mu_{\theta} = \mu_{\theta}(\tilde{\tau}^{i}, i, y)\) for brevity.
We approximate \(\log p(\mathcal{C} \mid \tilde{\tau}^{i-1})\) using a first-order Taylor expansion around the mean \(\tilde{\tau}^{i-1} = \mu_{\theta}\):
\begin{align}
  \log p(\mathcal{C} \!\mid\! \tilde{\tau}^{i-1}) 
  &\!\approx\! \log p(\mathcal{C} \mid \tilde{\tau}^{i-1}) \big|_{\tilde{\tau}^{i-1} = \mu_{\theta}} \!+\! (\tilde{\tau}^{i-1} - \mu_{\theta}) \cdot g \nonumber \\
  &\approx c_2 + (\tilde{\tau}^{i-1} - \mu_{\theta}) \cdot g,
\end{align}
where $c_2$ is the constant value at the mean, and $g = \nabla_{\tilde{\tau}^{i-1}} \log p(\mathcal{C} \mid \tilde{\tau}^{i-1}) \big|_{\tilde{\tau}^{i-1} = \mu_{\theta}}$ is the gradient of the safety log-probability evaluated at $\mu_{\theta}$. This linear approximation is valid locally and enables gradient-based guidance.
Combining the approximations, the conditional log-probability becomes:
\begin{minipage}{\linewidth}
\begin{align}
\label{eq:classifer_guidance_proof}
  &\log p_{\theta}(\tilde{\tau}^{i-1} \mid \tilde{\tau}^i, y, \mathcal{C})   \nonumber \\
  & \!\approx \!-\frac{1}{2\beta_i} \!\left\| \tilde{\tau}^{i-1} \!-\! \left( \mu_{\theta} \!+\! \frac{\beta_i}{\mathcal{T}} \nabla_{\tilde{\tau}^{i-1}} h(\tilde{\tau}^{i-1}) \bigg|_{\tilde{\tau}^{i-1} = \mu_{\theta}} \! \right) \! \right\|^2 \!+\! c_3 \nonumber \\
  & \approx -\frac{1}{2\beta_i} \|\tilde{\tau}^{i-1} - (\mu_{\theta} + \beta_i g)\|^2 + c_3,
\end{align}
\end{minipage}
where $c_3$ represents all constant terms.
This approximation in \eqref{eq:classifer_guidance_proof} reveals that the conditional distribution remains Gaussian, with its mean shifted in the direction of the safety constraint's gradient. 
Ultimately, since we rely on control actions to govern the robot's behavior, we apply this guidance by computing the gradients of the CBF and CLF conditions (as detailed in Section~\ref{subsec:gradient_cbf_clf}) with respect to the control input, rather than the entire trajectory. CBFs, as detailed in Section~\ref{subsec:CBF}, facilitate guiding the reverse diffusion process to produce trajectories that avoid collisions with static or dynamic obstacles. Similarly, CLFs can be utilized to steer the reverse diffusion process toward the goal.
\subsection{CBF and CLF Guidance Design}
\label{subsec:gradient_cbf_clf}

Since we consider both static and dynamic obstacles, their dynamics affect the CBF-based safety and CLF-based stability conditions. Therefore, we augment the nonlinear system model \eqref{eq:affine-dynamics_c} with the autonomous dynamics of these obstacles:
\begin{equation}
\label{eq:obst_dyn}
\dot{\mb x }_j = \mb{f}_j(\mb x_j), \ j = 1, \dots, N_J,
\end{equation}
where ${\mb x_j \!\in\! \R^{n_j}}$. We assume the future obstacle states $\mb{x}_{j,t+1}, \dots, \mb{x}_{j,t+H}$ over the planning horizon $H$ are known to the planner. Then, we define an augmented system dynamics formed by \eqref{eq:affine-dynamics_c} and \eqref{eq:obst_dyn} as
\begin{equation}
\label{eq:sys_aug}
\underbrace{\begin{bmatrix}
\dot{\mb x } \\ 
\dot{\mb x }_1 \\ 
\cdot \\
\dot{\mb x }_j \\
\cdot \\
\dot{\mb x }_{N_J}
\end{bmatrix}}_{\triangleq \dot{\bar{\mb x}}}
=
\underbrace{\begin{bmatrix}
\mb{f}(\mb{x}) \\ \mb{f}_1(\mb x_1) \\ 
\cdot \\
\mb{f}_j(\mb x_j) \\
\cdot \\
\mb{f}_{N_J}(\mb x_{N_J})
\end{bmatrix}}_{\triangleq \bar{\mb f}(\bar{\mb x})}  +
\underbrace{\begin{bmatrix}
\mb{g}(\mb{x}) \\ 
\bf 0  \\ 
\cdot \\
\bf 0 \\ 
\cdot \\
\bf 0 
\end{bmatrix}}_{\triangleq \tilde{\mb g}(\bar{\mb x})} \mb u,
\end{equation}
where ${\bar{\mb x} \!\in\! \R^{n + n_j \times N_J}}$ is the augmented state. We represent discrete-time dynamics as $\bar{\mb{F}}(\bar{\mb{x}}_k, \!\mb{u}_k )$.

Next, we consider the collision avoidance between the robot and the $j$-th obstacle, which has a radius ${r_j}$:
\begin{equation}
    h_j(\bar{\mb x}) = \|\mb{x} - \mb{x}_j\|^2 - r_j^2.
\end{equation}
Our goal is to satisfy all constraints simultaneously; therefore, we define a unified safety function using the $\min$ function: 
\begin{equation}
 h(\bar{\mb x}) =  \min_{j \in \{1,\ldots, N_J\}} \left [ h_j(\bar{\mb x}) \right ], 
\end{equation}
which is non-smooth. To construct a smooth function representing the $\min$ function, we can employ a smooth under-approximation given by (with  ${\lambda \!\geq\! 0}$):
\begin{equation}
\label{eq:softmin}
h(\bar{\mb x}) = -\dfrac{1}{\lambda} \ln \Big(\sum_{j=1}^{N_J} {\rm e}^{-\lambda h_j(\bar{\mb x})} \Big).
\end{equation}
From Theorem~\ref{thm: cbf}, we have a formal safety guarantee with respect to $h$ if the discrete-time CBF condition:
\begin{equation}
0 \leq h(\bar{\mb{F}}(\bar{\mb{x}}_k, \!\mb{u}_k )) - h(\bar{\mb x}_k) + \alpha (h(\bar{\mb x}_k))   \triangleq \psi_{h}(\bar{\mb{x}}_k, \mb{u}_k)
\end{equation}
is satisfied for all sampled $\bar{\mb x}_k$ via control input ${\mb u}_k$. 

Therefore, the collision avoidance safety \eqref{eq:softmin} is guaranteed if the safety condition, $\psi_{h}$, remains positive for every control action throughout the entire sequence. To actively enforce this using the guidance mechanism shown in \eqref{eq:classifer_guidance_proof}, we utilize the gradient of the safety condition with respect to the control input:
\begin{equation}
\nabla_{\mb{u}} \psi_{h}(\bar{\mb{x}}, \mb{u})
=
\Big( \frac{\partial \bar{\mb{F}} (\bar{\mb{x}}, \mb{u})}{\partial \mb{u}} \Big)^{\top}
\nabla_{\bar{\mb{x}}} h \! \left( \bar{\mb{F}}(\bar{\mb{x}}, \mb{u}) \right),
\label{eq:psi_grad_u}
\end{equation}
which points in the direction of steepest ascent of $\psi_h$ with respect to $\mb{u}$, indicating how to modify the control input to improve satisfaction of the discrete-time CBF condition \cite{mizuta2024cobl}. Under the Euler discretization, $\frac{\partial \bar{\mb{F}}}{\partial \mb{u}} = T \tilde{\mb{g}}(\bar{\mb{x}})$ depends only on $\bar{\mb{x}}$. Nevertheless, $\nabla_{\mb{u}}\psi_h(\bar{\mb{x}},\mb{u})$ generally depends on $\mb{u}$ through the evaluation of $\nabla_{\bar{\mb{x}}}h$ at $\bar{\mb{F}}(\bar{\mb{x}},\mb{u})$. By integrating this gradient as a guidance signal within each denoising step, evaluated at the current denoising iterate, we steer the diffusion model toward synthesizing trajectories directed toward the safe set.

Similar to how a CBF enforces safety by repelling the denoising steps from unsafe regions, a CLF can enforce stability by attracting the denoising steps toward a desired goal state. To guide the reverse diffusion process toward the goal state $g$, we define a CLF:
\begin{equation}
V(\bar{\mb x}) = \|\bar{\mb x} - g\|^2.
\end{equation}
The condition ensures exponential stability of $g$ is given by
\begin{align}
   0 \!\geq\! V(\bar{\mb{F}}(\bar{\mb x}_k, \!\mb{u}_k )) - V( \bar{\mb x}_k) + \sigma  V( \bar{\mb x}_k) \!\triangleq\! \Phi_{V}(\bar{\mb{x}}_k, \mb{u}_k),
\end{align}
The gradient of $\Phi_{V}$ with respect to $u$: 
\begin{equation}
\nabla_{\mb{u}} \Phi_{V}(\bar{\mb{x}}, \mb{u})
=
\Big( \frac{\partial \bar{\mb{F}}(\bar{\mb{x}}, \mb{u})}{\partial \mb{u}} \Big)^{\top}
\nabla_{\bar{\mb{x}}} V\! \left(\bar{\mb{F}}(\bar{\mb{x}}, \mb{u}) \right),
\label{eq:phiV_grad_u}
\end{equation}

which indicates the direction in which $\mb{u}$ should be selected to most effectively decrease the CLF value, thereby enabling convergence to the goal state $g$.

By integrating gradients from both CBFs and CLFs into the denoising process as
\begin{equation}
  p_{\theta}(\tilde{\tau}^{i - 1} \mid \tilde{\tau}^{i}, y)= \mathcal{N}(\mu_{\theta} + \lambda_c \nabla_{\mathbf{u}}\psi_{h} - \lambda_l \nabla_{\mathbf{u}}\Phi_{V}, \beta_i \mathbf{I}),
\end{equation}
the diffusion model generates trajectories that avoid obstacles while advancing toward the goal, thereby reducing the risk of stagnation in safe yet suboptimal regions. Nevertheless, conflicts may arise when the CBF and CLF gradients oppose one another. To address this, we employ an adaptive weight ${\lambda_c \!\geq\! 0}$ for the CBF constraint, which increases proportionally as the distance to obstacles decreases, while maintaining a fixed weight ${\lambda_l \!\geq\! 0}$ for the CLF constraint.
Note that we set both ${\lambda_c, \lambda_l \!=\! 0}$ during the first few denoising steps, as the initial noisy trajectories cannot be expected to satisfy safety constraints.

Although CBFs and CLFs guide the generated trajectories by steering them away from obstacles and toward desired goals, respectively, this guidance serves as a soft directive rather than a hard constraint, steering the model toward approximately feasible regions. As a result, safety violations may still arise in the generated trajectories.
To mitigate this issue, we utilize an MPC as a corrective refinement to enhance the trajectories, guaranteeing compliance with both dynamic and safety constraints. We can formulate the MPC as

\begin{equation}
\label{eq:mpc}
\begin{aligned}
    \min_{\mathbf{u}_{0:H}, \bar{\mathbf{x}}_{0:H}} \quad & 
    \sum_{t=0}^H \frac{1}{2} ({{\tau^i}})^\top Q ({{\tau^i}}) - (\tilde{\tau}^i)^\top Q ({{\tau^i}}) \\
    \text{s.t} \quad &  h(\bar{\mb x}_k) \geq 0 \\
    & \bar{{\mb x }}_{k+1} = \bar{\mb{F}}(\bar{\mb{x}}_k, \mb{u}_k ) \\
    & \mathbf{\tau}^{i-1}_{0} = \tilde{\tau}^{i-1}
\end{aligned}
\end{equation}
where ${\tau^i \triangleq (\bar{\mb x}_{0}^i, \mb u_{0}^i \dots, \bar{\mb x}_{H}^i)}$ and $\tilde{\tau}^{i-1}$ is the warm-start trajectory from the CBF- and CLF-guided diffusion process, providing a probabilistically safe initialization for the MPC. This initialization enables the MPC to converge more rapidly to feasible, goal-oriented solutions compared to unguided cases, where it may fail to identify valid trajectories within allocated iteration limits or become trapped in safe yet suboptimal regions. 
In the MPC objective function given in \eqref{eq:mpc}, the first term, \(\frac{1}{2} (\tau^i)^\top Q (\tau^i)\), serves as a regularization term that prevents the optimization variables from growing excessively large. The second term, \(-(\tilde{\tau}^i)^\top Q (\tau^i)\), encourages the optimization variables to remain close to the output of the guided diffusion model, ensuring that the resulting trajectories do not deviate significantly from the distribution learned by the diffusion neural network.
Similar to the approach in DPCC \cite{romer2025dpcc}, our framework employs an iterative projection scheme that tightly integrates MPC into the reverse diffusion process. At each denoising iteration, \footnotetext{Since the data contain more noise in the early denoising iterations, guidance and MPC projection are excluded for the first few steps.} the MPC projects the probabilistically safe trajectory $\tilde{\tau}^{i-1}$ onto a safe and dynamically feasible set. This is achieved by using $\tilde{\tau}^{i-1}$ to both initialize the MPC's solver (as a warm start) and penalize deviations from this trajectory in the cost function. The resulting corrected trajectory then initializes the subsequent denoising step. In this refinement loop, MPC serves as a dedicated projection operator, rigorously enforcing constraints throughout the generation process.
Although only the final denoised trajectory is executed, enforcing feasibility during intermediate steps is important due to the sequential (Markovian) nature of reverse diffusion: each iterate initializes the subsequent denoising step and, in our method, also warm-starts the MPC projection. Consequently, a highly unsafe or dynamically infeasible intermediate trajectory can propagate errors forward by biasing subsequent denoising updates or by preventing the MPC from converging within a fixed computational budget, ultimately degrading the final plan. Following DPCC \cite{romer2025dpcc}, we therefore apply the projection operator during denoising, but only after an initial burn-in phase (once the sample leaves the pure-noise regime); from that point onward, the projected feasible iterate is fed back to the diffusion model to keep the remaining denoising iterations within the basin of attraction of the constrained optimizer.
The overall framework is visualized in Fig.~\ref{fig:model_architecture}.

\begin{figure}
    \centering
    \vspace{2.2mm} \includegraphics[width=1\linewidth]{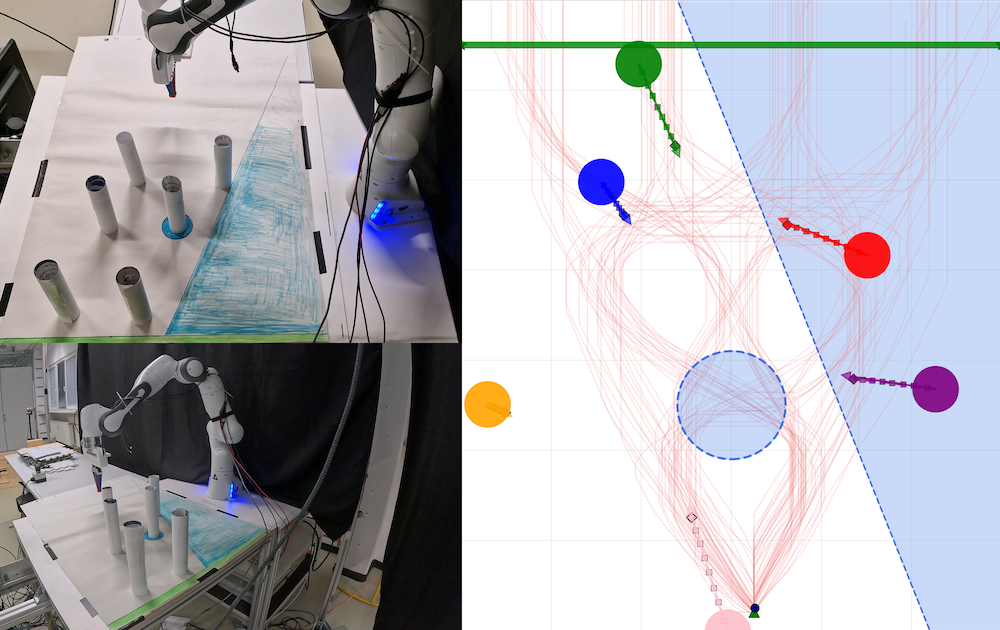}
    \caption{Experimental setups for static (left) and dynamic (right) obstacle avoidance scenarios. The left panel shows the physical setup for our Sim2Real experiments, featuring static obstacles (white cylinders), a designated safe area (blue), and a goal line (green). The right panel displays the simulation environment with dynamic obstacles (colored circles), whose future paths are indicated by arrows. The faint red lines represent the numerous expert trajectories trained by our framework to safely navigate the environment, while the light blue area indicates a safety constraint.}
    \label{fig:sim2real}
\end{figure}

\begin{algorithm}[tb]
  \caption{\algname with CBF/CLF Guidance}
  \label{alg:d_safempc_with_guidance}
  \begin{algorithmic}[1]
  \State \textbf{Input:} Guidance weights $\lambda_c, \lambda_l$
  \State Initialize trajectory ${\tau}^N \sim \mathcal{N}(0, I)$ \label{alg1:initialize}
      \For{$i = N \dots 1$} \label{alg1:iterative_start}
          \State $z \sim \mathcal{N}(0, I) \;\text{if } i > 1, \;\text{else } z = 0$
          \State $\mu_\theta = \frac{1}{\sqrt{\alpha_i}} \left( {\tau}^{i} - \frac{\beta_i}{\sqrt{1 - \bar{\alpha}_i}} \, \epsilon_\theta({\tau}^{i}, i, y) \right)$ \label{alg1:iterative_predict_mean}
          \State \textit{Denoise with CBF/CLF guidance}
          \State $\hat{\mu} \leftarrow \mu_\theta + \lambda_c \nabla_{\mathbf{u}}\psi_{h} - \lambda_l \nabla_{\mathbf{u}}\Phi_{V}$ \label{alg1:iterative_gradients}
          \State $\tilde{\tau}^{i-1} \leftarrow \! \hat{\mu} \!+ z  \sigma_i$ \Comment{\small Sample using reparameterization trick} \label{alg1:sampling_new_trajectory}
          \If{$i \leq N/2$}
          \State $\tau^{i-1} \leftarrow \texttt{MPC-Project}(\tilde{\tau}^{i-1})$ \label{alg1:iterative_mpc_project} \Comment{Eq.\eqref{eq:mpc}}
          \Else
             \State $\tau^{i-1} \leftarrow \tilde{\tau}^{i-1}$
          \EndIf
      \EndFor \label{alg1:iterative_end}
  \State \textbf{return} $\tau^0$
  \end{algorithmic}
  \end{algorithm}

Algorithm~\ref{alg:d_safempc_with_guidance} outlines the \algname framework. The framework begins by initializing a trajectory with pure Gaussian noise (line~\ref{alg1:initialize}). In Iterative mode (lines~\ref{alg1:iterative_start}--\ref{alg1:iterative_end}), it employs a step-by-step reverse diffusion process from diffusion step $N$ down to 1. In each iteration, the following steps occur:

\begin{itemize}
\item Guidance: The predicted mean of the reverse process, $\mu$, (line~\ref{alg1:iterative_predict_mean}) is adjusted using gradients from CBFs and CLFs (line~\ref{alg1:iterative_gradients}). This critical step steers the trajectory generation toward safer, goal-oriented regions while avoiding obstacles.

\item Sampling: A new, less noisy trajectory, $\tilde{\tau}^{i-1}$, is sampled from a distribution centered at the guided mean, $\hat{\mu}$ (line~\ref{alg1:sampling_new_trajectory}).

\item Projection: Once the trajectory leaves the high-noise regime ($i \leq N/2$)\footnote{To ensure a fair comparison, we adopt the same burn-in strategy and hyperparameters as the DPCC baseline \cite{romer2025dpcc}.}, an MPC projects it onto a safe and dynamically feasible set (line~\ref{alg1:iterative_mpc_project}) as expressed in \eqref{eq:mpc}. The resulting corrected trajectory, $\tau^{i-1}$, is then used to initialize the subsequent denoising step, ensuring that safety and system constraints are enforced throughout the entire generation process. 
\end{itemize}
\begin{table*}[t]
\vspace{2.2mm}
  \caption{Planning performance in static and dynamic environments, showing average success and safe-success rates.}
  \label{tab:quantitative-comparison}
  \centering
  \begin{adjustbox}{width=\textwidth}
    \setlength{\tabcolsep}{4pt}
    \renewcommand{\arraystretch}{1.15}
    \begin{tabular}{l *{8}{U}}
      \toprule
      & \multicolumn{2}{c}{\textbf{Static Env}}
      & \multicolumn{2}{c}{\textbf{Dynamic Env 1}}
      & \multicolumn{2}{c}{\textbf{Dynamic Env 2}}
      & \multicolumn{2}{c}{\textbf{Dynamic Env 3}} \\
      \cmidrule(lr){2-3}\cmidrule(lr){4-5}\cmidrule(lr){6-7}\cmidrule(lr){8-9}
      \textbf{Planner}
        & \multicolumn{1}{c}{\makecell{\textbf{Safety-Compliant} \\\textbf{Goal Reached Rate}}}
        & \multicolumn{1}{c}{\makecell{\textbf{Goal Reached}\\\textbf{Rate}}}
        & \multicolumn{1}{c}{\makecell{\textbf{Safety-Compliant} \\\textbf{Goal Reached Rate}}}
        & \multicolumn{1}{c}{\makecell{\textbf{Goal Reached}\\\textbf{Rate}}}
        & \multicolumn{1}{c}{\makecell{\textbf{Safety-Compliant} \\\textbf{Goal Reached Rate}}}
        & \multicolumn{1}{c}{\makecell{\textbf{Goal Reached}\\\textbf{Rate}}}
        & \multicolumn{1}{c}{\makecell{\textbf{Safety-Compliant} \\\textbf{Goal Reached Rate}}}
        & \multicolumn{1}{c}{\makecell{\textbf{Goal Reached}\\\textbf{Rate}}} \\
      \midrule
      \textbf{\algname (ours)}
        & 0.800 \pm 0.332 
        & 0.800 \pm 0.332 
        & \cellcolor{blond} 1.000 \pm 0.000 
        & \cellcolor{blond} 1.000 \pm 0.000 
        & \cellcolor{blond} 0.875 \pm 0.268 
        & 0.875 \pm 0.268 
        & \cellcolor{blond} 0.925 \pm 0.179 
        & \cellcolor{blond} 0.925 \pm 0.179 
        \\
      \textbf{CoBL}
        & 0.050 \pm 0.150 
        & \cellcolor{blond} 1.000 \pm 0.000 
        & 0.100 \pm 0.300 
        & 0.850 \pm 0.320 
        & 0.775 \pm 0.370 
        & \cellcolor{blond} 0.925 \pm 0.238 
        & 0.800 \pm 0.332 
        & 0.850 \pm 0.320 
        \\
      \textbf{Diffuser}
        & 0.050 \pm 0.150 
        & 0.650 \pm 0.450 
        & 0.200 \pm 0.332 
        & 0.400 \pm 0.436 
        & 0.175 \pm 0.327 
        & 0.325 \pm 0.426 
        & 0.325 \pm 0.426 
        & 0.450 \pm 0.415 
        \\
      \textbf{DPCC-C}
        & \cellcolor{blond} 0.900 \pm 0.300 
        & 0.900 \pm 0.300 
        & 0.750 \pm 0.335 
        & 0.850 \pm 0.229 
        & 0.450 \pm 0.415 
        & 0.600 \pm 0.339 
        & 0.425 \pm 0.327 
        & 0.425 \pm 0.327 
        \\
      \textbf{DPCC-C Tightened}
        & 0.700 \pm 0.400 
        & 0.700 \pm 0.400 
        & 0.700 \pm 0.332 
        & 0.750 \pm 0.250 
        & 0.575 \pm 0.396 
        & 0.575 \pm 0.396 
        & 0.650 \pm 0.391 
        & 0.650 \pm 0.391 
        \\
      \textbf{DPCC-R}
        & 0.350 \pm 0.391 
        & 0.500 \pm 0.387 
        & 0.450 \pm 0.350 
        & 0.650 \pm 0.391 
        & 0.275 \pm 0.295 
        & 0.325 \pm 0.327 
        & 0.675 \pm 0.286 
        & 0.825 \pm 0.238 
        \\
      \textbf{DPCC-R Tightened}
        & 0.700 \pm 0.332 
        & 0.700 \pm 0.332 
        & 0.800 \pm 0.332 
        & 0.900 \pm 0.300 
        & 0.600 \pm 0.339 
        & 0.600 \pm 0.339 
        & 0.750 \pm 0.250 
        & 0.750 \pm 0.250 
        \\
      \textbf{DPCC-T}
        & 0.800 \pm 0.332 
        & 0.950 \pm 0.150 
        & 0.350 \pm 0.320 
        & 0.500 \pm 0.316 
        & 0.050 \pm 0.150 
        & 0.050 \pm 0.150 
        & 0.275 \pm 0.334 
        & 0.300 \pm 0.367 
        \\
      \textbf{DPCC-T Tightened}
        & 0.850 \pm 0.320 
        & 0.850 \pm 0.320 
        & 0.450 \pm 0.350 
        & 0.450 \pm 0.350 
        & 0.125 \pm 0.217 
        & 0.125 \pm 0.217 
        & 0.350 \pm 0.421 
        & 0.350 \pm 0.421 
        \\
      \textbf{Guidance}
        & 0.400 \pm 0.490 
        & 0.400 \pm 0.490 
        & 0.400 \pm 0.490 
        & 0.400 \pm 0.490 
        & 0.525 \pm 0.487 
        & 0.525 \pm 0.487 
        & 0.525 \pm 0.487 
        & 0.525 \pm 0.487 
        \\
      \textbf{Guidance Tightened}
        & 0.400 \pm 0.490 
        & 0.400 \pm 0.490 
        & 0.400 \pm 0.490 
        & 0.400 \pm 0.490 
        & 0.525 \pm 0.487 
        & 0.525 \pm 0.487 
        & 0.525 \pm 0.487 
        & 0.525 \pm 0.487 
        \\        
      \textbf{Model-Free}
        & 0.000 \pm 0.000 
        & 0.550 \pm 0.415 
        & 0.350 \pm 0.450 
        & 0.450 \pm 0.472 
        & 0.225 \pm 0.370 
        & 0.300 \pm 0.400 
        & 0.350 \pm 0.421 
        & 0.450 \pm 0.415 
        \\
      \textbf{Model-Free Tightened}
        & 0.000 \pm 0.000 
        & 0.600 \pm 0.374 
        & 0.350 \pm 0.450 
        & 0.500 \pm 0.447 
        & 0.275 \pm 0.402 
        & 0.425 \pm 0.396 
        & 0.375 \pm 0.415 
        & 0.425 \pm 0.426 
        \\
      \textbf{Post-Processing}
        & 0.550 \pm 0.350 
        & 0.750 \pm 0.335 
        & 0.550 \pm 0.350 
        & 0.600 \pm 0.300 
        & 0.400 \pm 0.374 
        & 0.425 \pm 0.396 
        & 0.550 \pm 0.350 
        & 0.725 \pm 0.295 
        \\
      \textbf{Post-Processing Tightened}
        & 0.700 \pm 0.332 
        & 0.700 \pm 0.332 
        & 0.750 \pm 0.403 
        & 0.750 \pm 0.403 
        & 0.600 \pm 0.300 
        & 0.600 \pm 0.300 
        & 0.700 \pm 0.292 
        & 0.700 \pm 0.292 
        \\
      \bottomrule
    \end{tabular}
  \end{adjustbox}
  \vspace{-3mm}
\end{table*}

\begin{figure}
    \centering
   \includegraphics[width=0.9\linewidth]{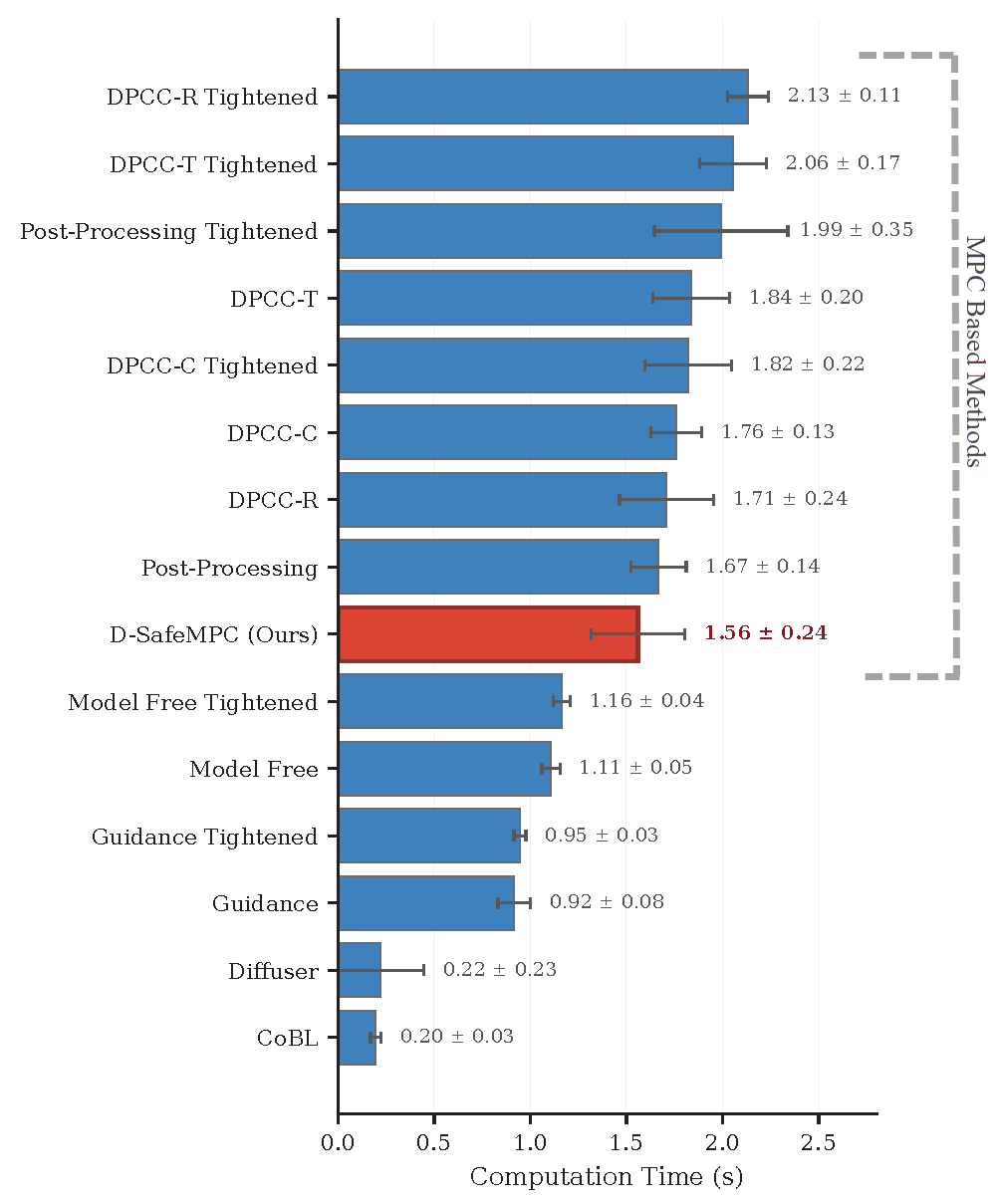}
    \caption{A comparison of the average time (in seconds) each planner requires to generate a control action in the static obstacle environment.}
    \label{fig:computation_time_per_method}
     \vspace{-4mm}
\end{figure}

\section{Experiments}
\label{sec:imple}
In our simulation, we create realistic dynamic environments for robotic path planning by modeling obstacle motion with parametric trajectories. Obstacles are categorized as either static, with fixed positions, or dynamic, following circular orbits with a defined angular velocity.

For each dynamic obstacle, we generate an ideal reference trajectory in $\mathbb{R}^2$, denoted by $p_{\text{ref}}(t)$. The position reference at time $t$ is defined as $x_{\text{ref}}(t) = x_c + r \cos(\omega t + \phi)$ and $y_{\text{ref}}(t) = y_c + r \sin(\omega t + \phi)$
where ($x_c, y_c$) is the center of the circular path, $r$ is the radius, $\omega$ is the angular velocity, and $\phi$ is the initial phase offset. For static obstacles, the reference trajectory is a fixed point, which is achieved by setting the radius and angular velocity to zero ($r=0, \omega=0$).

To create a diverse set of scenarios, the parameters for each obstacle's trajectory—radius, angular velocity, and phase offset—are sampled from uniform distributions within specified bounds. The actual obstacle trajectories are simulated using a proportional-derivative (PD) controller to track the reference path. The system's dynamics are modeled as a velocity-controlled point mass, with the control input determined by the PD control law. The simulation uses forward Euler integration to update the obstacle's position at each timestep. In all experiments, we use $N=20$ diffusion steps; consequently, the burn-in threshold in Algorithm~\ref{alg:d_safempc_with_guidance} corresponds to $i \leq N/2 = 10$. 

We evaluate all planners across 10 random seeds using two metrics (Table~\ref{tab:quantitative-comparison})\footnote{Each planner is tested with two variants per seed, totaling 20 evaluations.}. The \textit{Goal Reach Rate} measures the proportion of such collision-free goal-reaching trials. The \textit{Safety-Compliant Goal Reached Rate} additionally requires satisfaction of all safety constraints throughout the trajectory.

Our proposed framework, \algname, demonstrates consistently superior performance over multiple baselines across both static and dynamic environments. In the static obstacle scenario, while CoBL achieves marginally higher success rates and DPCC-C attains better safe-success rates, our method remains highly competitive, with \algname yielding 80\% success and safe-success. Notably, \algname excels in dynamic obstacle scenarios, where it achieves the highest or near-highest success and safe-success rates, underscoring its robustness in enforcing safety constraints under challenging conditions.
In addition to task performance, we evaluated the computational efficiency of each planner. Figure~\ref{fig:computation_time_per_method} illustrates the average computation time per decision step for each method in the static environment. Our proposed framework, \algname, achieves a favorable balance between its high safe-success rates and computational overhead. Notably, \algname is computationally more efficient than other MPC-based models, such as the DPCC variants and Post-Processing. We attribute this improvement to the CBF and CLF guidance, which provides the MPC with better-initialized trajectories, leading to faster convergence. While methods without an MPC projection step, such as Diffuser, are significantly faster, their speed comes at the cost of lower safety and reliability, as reflected in their poor safe-success rates.

\subsection{Sim-to-Real Experiments}
We have validated our framework with a sim-to-real experiment, using the physical (static obstacles) and simulated (dynamic obstacles) setups shown in Fig.~\ref{fig:sim2real}.
The diffusion model, trained entirely in simulation, has been deployed on a physical Franka robot manipulator. Control and state feedback have been handled using the Franky library\footnote{\href{https://github.com/TimSchneider42/franky}{https://github.com/TimSchneider42/franky}}. During this sim-to-real experiment, we have scaled the dynamic parameters of the real robot by setting the relative dynamics factor to 0.05 in order to minimize controller errors and ensure precise positioning, which is crucial for successfully avoiding obstacles and unsafe regions. In the experiment, the robot has executed a trajectory guided by CBFs and CLFs, with final optimization performed by the MPC. We have demonstrated the real-world performance of our proposed \algname and DPCC baseline in a sim-to-real experiment.
\section{Conclusion}
\label{sec:conc}
In this work, we introduce \algname, a novel framework that addresses the critical challenge of enforcing safety and dynamic feasibility in diffusion-based trajectory planning. By guiding the reverse diffusion process with discrete-time CBF and CLF, our approach steers trajectory generation toward safe, goal-directed regions.
This guidance provides high-quality warm starts for a subsequent MPC.
Our framework's effectiveness is demonstrated through an iterative-projection scheme where the MPC refines the trajectory at each denoising step, providing robust refinement and enforcing strict safety and dynamic constraints. Through extensive simulations in environments with both static and dynamic obstacles, alongside a sim-to-real experiment on a physical Franka manipulator, we have shown that \algname significantly improves safety, task success rates, and planning efficiency compared to state-of-the-art baselines. 

Despite these advancements, our framework has limitations. It may not scale well to high-dimensional state spaces (e.g., visual inputs), as MPC optimization becomes computationally prohibitive. Additionally, positioning inaccuracies in early time steps can propagate recursively, since the diffusion model conditions trajectory generation on the current state, potentially destabilizing subsequent predictions.

Future work will focus on extending this framework to systems with more complex, higher-dimensional state spaces while addressing scalability and error-propagation challenges.
\end{spacing}

\vspace{-1mm}
\begin{spacing}{0.96}
\bibliographystyle{IEEEtran}
\bibliography{references}

@inproceedings{romer2025dpcc,
  title = 	 {Diffusion Predictive Control with Constraints},
  author =       {R{\"o}mer, Ralf and Rohr, Alexander von and Schoellig, Angela},
  booktitle = 	 {Proceedings of the 7th Annual Learning for Dynamics \& Control Conference},
  pages = 	 {791--803},
  year = 	 {2025},
  volume = 	 {283},
  publisher =    {PMLR},
}

@article{jia2024towards,
	title        = {Towards diverse behaviors: A benchmark for imitation learning with human demonstrations},
	author       = {Jia, Xiaogang and Blessing, Denis and Jiang, Xinkai and Reuss, Moritz and Donat, Atalay and Lioutikov, Rudolf and Neumann, Gerhard},
	year         = 2024,
	journal      = {arXiv preprint arXiv:2402.14606}
}

@inproceedings{janner2022planning,
  title     = {Planning with Diffusion for Flexible Behavior Synthesis},
  author    = {Janner, Michael and Du, Yilun and Tenenbaum, Joshua and Levine, Sergey},
  booktitle = {ICML},
  year      = {2022}
}

@article{chi2025diffusionpolicy,
  author  = {Chi, Cheng and Xu, Zhenjia and Feng, Siyuan and Cousineau, Eric and Du, Yilun and Burchfiel, Benjamin and Tedrake, Russ and Song, Shuran},
  title   = {Diffusion Policy: Visuomotor Policy Learning via Action Diffusion},
  journal = {The International Journal of Robotics Research},
  year    = {2025}
}

@inproceedings{carvalho2023motion,
	title        = {Motion planning diffusion: Learning and planning of robot motions with diffusion models},
	author       = {Carvalho, Joao and Le, An T and Baierl, Mark and Koert, Dorothea and Peters, Jan},
	year         = 2023,
	booktitle    = {2023 IEEE/RSJ International Conference on Intelligent Robots and Systems (IROS)},
	pages        = {1916--1923},
	organization = {IEEE}
}

@inproceedings{xiao2023safediffuser,
	title        = {Safediffuser: Safe planning with diffusion probabilistic models},
	author       = {Xiao, Wei and Wang, Tsun-Hsuan and Gan, Chuang and Hasani, Ramin and Lechner, Mathias and Rus, Daniela},
	year         = 2023,
	booktitle    = {The Thirteenth International Conference on Learning Representations}
}

@article{zhou2024diffusion,
	title        = {Diffusion model predictive control},
	author       = {Zhou, Guangyao and Swaminathan, Sivaramakrishnan and Raju, Rajkumar Vasudeva and Guntupalli, J Swaroop and Lehrach, Wolfgang and Ortiz, Joseph and Dedieu, Antoine and L{\'a}zaro-Gredilla, Miguel and Murphy, Kevin},
	year         = 2024,
	journal      = {arXiv preprint arXiv:2410.05364}
}

@inproceedings{NEURIPS2021_49ad23d1,
	title        = {Diffusion Models Beat {GAN}s on Image Synthesis},
	author       = {Dhariwal, Prafulla and Nichol, Alexander},
	year         = 2021,
	booktitle    = {Advances in Neural Information Processing Systems},
	publisher    = {Curran Associates, Inc.},
	volume       = 34,
	pages        = {8780--8794},
}

@inproceedings{AmesCooganEtAl2019,
	title        = {Control barrier functions: Theory and applications},
	author       = {Ames, A.~D. and Coogan, S. and Egerstedt, M. and Notomista, G. and Sreenath, K. and Tabuada, P.},
	year         = 2019,
	booktitle    = {2019 18th European Control Conference (ECC)}
}

@inproceedings{NEURIPS2020_4c5bcfec,
	title        = {Denoising Diffusion Probabilistic Models},
	author       = {Ho, Jonathan and Jain, Ajay and Abbeel, Pieter},
	year         = 2020,
	booktitle    = {{Advances in Neural Information Processing Systems}},
	publisher    = {Curran Associates, Inc.},
	volume       = 33,
	pages        = {6840--6851},
}

@article{Blei_2017,
	title        = {Variational Inference: A Review for Statisticians},
	author       = {Blei, David M. and Kucukelbir, Alp and McAuliffe, Jon D.},
	year         = 2017,
	month        = apr,
	journal      = {Journal of the American Statistical Association},
	publisher    = {Informa UK Limited},
	volume       = 112,
	number       = 518,
	pages        = {859–877},
	doi          = {10.1080/01621459.2017.1285773},
	issn         = {1537-274X},
}

@article{ames2017control,
  author={A. D. Ames and X. Xu and J. W. Grizzle and P. Tabuada},
  journal={IEEE Transactions on Automatic Control},
  title={Control barrier function based quadratic programs for safety critical systems},
  year={2017},
  volume={62},
  number={8},
  pages={3861--3876},
  publisher={IEEE}
}

@inproceedings{kurtz2025equality,
  title={Equality constrained diffusion for direct trajectory optimization},
  author={Kurtz, Vince and Burdick, Joel W},
  booktitle={American Control Conference (ACC)},
  pages={535--540},
  year={2025},
  organization={IEEE}
}

@article{eren2017mpc,
author = {Eren, Utku and Prach, Anna and Ko\c{c}er, Ba\c{s}aran Bahad\i{}r and Rakovi\'{c}, Sa\v{s}a V. and Kayacan, Erdal and A\c{c}\i{}kme\c{s}e, Beh\c{c}et},
title = {Model Predictive Control in Aerospace Systems: Current State and Opportunities},
journal = {Journal of Guidance, Control, and Dynamics},
volume = {40},
number = {7},
pages = {1541-1566},
year = {2017},
doi = {10.2514/1.G002507},
}

@inproceedings{todorov2012mujoco,
  title={{MuJoCo: A physics engine for model-based control}},
  author={Todorov, Emanuel and Erez, Tom and Tassa, Yuval},
  booktitle={IEEE/RSJ International Conference on Intelligent Robots and Systems},
  pages={5026--5033},
  year={2012},
  organization={IEEE}
}

@inproceedings{mizuta2024cobl,
  title={{CoBL-Diffusion}: Diffusion-based conditional robot planning in dynamic environments using control barrier and {Lyapunov} functions},
  author={Mizuta, Kazuki and Leung, Karen},
  booktitle={2024 IEEE/RSJ International Conference on Intelligent Robots and Systems (IROS)},
  pages={13801--13808},
  year={2024},
  organization={IEEE}
}

@inproceedings{romer2024safe,
title={Safe Offline Reinforcement Learning using Trajectory-Level Diffusion Models},
author={Ralf R{\"o}mer and Lukas Brunke and Martin Schuck and Angela P. Schoellig},
booktitle={ICRA 2024 Workshop{\textemdash}Back to the Future: Robot Learning Going Probabilistic},
year={2024}
}

@article{giannone2023aligning,
  title={Aligning optimization trajectories with diffusion models for constrained design generation},
  author={Giannone, Giorgio and Srivastava, Akash and Winther, Ole and Ahmed, Faez},
  journal={Advances in neural information processing systems},
  volume={36},
  pages={51830--51861},
  year={2023}
}

@inproceedings{agrawal2017,
  title={Discrete control barrier functions for safety-critical control of discrete systems with application to bipedal robot navigation.},
  author={Agrawal, Ayush and Sreenath, Koushil},
  booktitle={Robotics: Science and Systems},
  volume={13},
  pages={1--10},
  year={2017},
  organization={Cambridge, MA, USA}
}

@inproceedings{ahmadi2019safe,
  title={Safe policy synthesis in multi-agent POMDPs via discrete-time barrier functions},
  author={Ahmadi, Mohamadreza and Singletary, Andrew and Burdick, Joel W and Ames, Aaron D},
  booktitle={2019 IEEE 58th Conference on Decision and Control (CDC)},
  pages={4797--4803},
  year={2019},
  organization={IEEE}
}

@inproceedings{nichol2021improved,
  title={Improved denoising diffusion probabilistic models},
  author={Nichol, Alexander Quinn and Dhariwal, Prafulla},
  booktitle={International conference on machine learning},
  pages={8162--8171},
  year={2021},
  organization={PMLR}
}

@article{carvalho2025motion,
  title={Motion planning diffusion: Learning and adapting robot motion planning with diffusion models},
  author={Carvalho, Jo{\~a}o and Le, An T and Kicki, Piotr and Koert, Dorothea and Peters, Jan},
  journal={IEEE Transactions on Robotics},
  year={2025},
  publisher={IEEE}
}

@INPROCEEDINGS{10406329,
  author={Amer, Abdelhakim and Mehndiratta, Mohit and le Fevre Sejersen, Jonas and Pham, Huy Xuan and Kayacan, Erdal},
  booktitle={21st International Conference on Advanced Robotics (ICAR)}, 
  title={Visual Tracking Nonlinear Model Predictive Control Method for Autonomous Wind Turbine Inspection}, 
  year={2023},
  volume={},
  number={},
  pages={431-438}}

@ARTICLE{10916556,
  author={Amer, Abdelhakim and Mehndiratta, Mohit and Brodskiy, Yury and Kayacan, Erdal},
  journal={IEEE Transactions on Control Systems Technology}, 
  title={Empowering Autonomous Underwater Vehicles Using Learning-Based Model Predictive Control With Dynamic Forgetting {G}aussian Processes}, 
  year={2025},
  volume={33},
  number={5},
  pages={1913-1920}}
\end{spacing}

\thispagestyle{plain}
\end{document}